\documentclass[10pt,twocolumn,letterpaper]{article}

\usepackage{iccv}
\usepackage{times}
\usepackage{epsfig}
\usepackage{graphicx}
\usepackage{amsmath}
\usepackage{amssymb}
\usepackage{graphicx}
\usepackage{amsmath}
\usepackage{amssymb}
\usepackage{booktabs}
\usepackage{times}
\usepackage{microtype}
\usepackage{epsfig}
\usepackage[table,xcdraw]{xcolor}
\usepackage{caption}
\usepackage{xcolor}
\usepackage{float}
\usepackage{placeins}
\usepackage{color, colortbl}
\usepackage{stfloats}
\usepackage{enumitem}
\usepackage{tabularx}
\usepackage{xstring}
\usepackage{multirow}
\usepackage{xspace}
\usepackage{url}
\usepackage{subcaption}
\usepackage{arydshln}
\usepackage{xcolor}
\usepackage{bbm}
\usepackage[hang,flushmargin]{footmisc}
\usepackage{pifont}%
\usepackage{lipsum}

\usepackage[accsupp]{axessibility}  

\newcommand{\et}[2]{${#1}^{\pm{#2}}$}

\newcommand{\etr}[2]{$\textcolor{black}{{#1}}^{\pm{#2}}$}
\newcommand{\etbb}[2]{$\textcolor{black}{{#1}}^{\pm{#2}}$}

\usepackage[pagebackref=true,breaklinks=true,letterpaper=true,colorlinks,bookmarks=false]{hyperref}
\usepackage[capitalize]{cleveref}

\crefname{section}{Sec.}{Secs.}
\Crefname{section}{Section}{Sections}
\Crefname{table}{Table}{Tables}
\crefname{table}{Tab.}{Tabs.}

\iccvfinalcopy 

\def\iccvPaperID{4089} 

\ificcvfinal\fi

\begin{document}

\title{Priority-Centric Human Motion Generation in Discrete Latent Space}

\author{
Hanyang Kong$^{1}$, Kehong Gong$^{1}$, Dongze Lian$^{1}$, Michael Bi Mi$^{2}$, Xinchao Wang$^{1}$\thanks{Corresponding author:
xinchao@nus.edu.sg}
\vspace{0.2cm}\\
$^1$National University of Singapore
\quad
$^2$Huawei International Pte Ltd\\
\texttt{\small \{hanyang.k, gongkehong\}@u.nus.edu}
\quad
\texttt{\small \{dongze, xinchao\}@nus.edu.sg}
}

\maketitle

\ificcvfinal\thispagestyle{empty}\fi

\begin{abstract}
   Text-to-motion generation is a formidable task, aiming to produce human motions that align with the input text while also adhering to human capabilities and physical laws. While there have been advancements in diffusion models, their application in discrete spaces remains underexplored. Current methods often overlook the varying significance of different motions, treating them uniformly. It is essential to recognize that not all motions hold the same relevance to a particular textual description. Some motions, being more salient and informative, should be given precedence during generation. In response, we introduce a Priority-Centric Motion Discrete Diffusion Model (M2DM), which utilizes a Transformer-based VQ-VAE to derive a concise, discrete motion representation, incorporating a global self-attention mechanism and a regularization term to counteract code collapse. We also present a motion discrete diffusion model that employs an innovative noise schedule, determined by the significance of each motion token within the entire motion sequence. This approach retains the most salient motions during the reverse diffusion process, leading to more semantically rich and varied motions. Additionally, we formulate two strategies to gauge the importance of motion tokens, drawing from both textual and visual indicators. Comprehensive experiments on the HumanML3D and KIT-ML datasets confirm that our model surpasses existing techniques in fidelity and diversity, particularly for intricate textual descriptions.
\end{abstract}

\section{Introduction}

\begin{figure*}[htbp] 
\centering 
\includegraphics[width=1.0\textwidth]{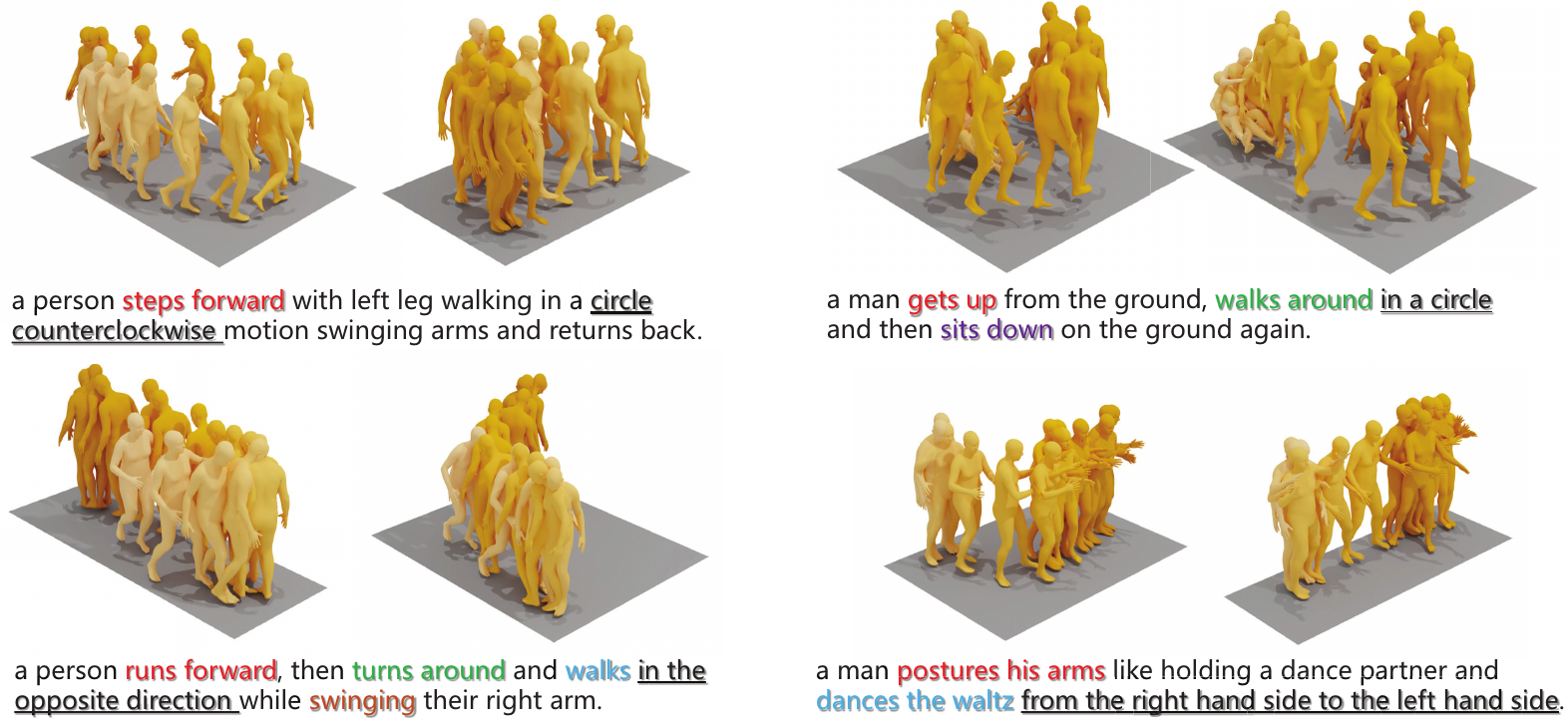} 
\caption{Our Priority-Centric Motion Discrete Diffusion Model (M2DM) generates diverse and precise human motions given complex textual descriptions. The color of human mesh goes from light to dark over time.}
\label{fig1}
\end{figure*}

Generating realistic and diverse 3D human motions under various conditions \eg, action labels \cite{petrovich2021action, guo2020action2motion}, natural language descriptions \cite{zhang2023t2m, guo2022tm2t, chen2022executing, zhang2022motiondiffuse}, and musical cues \cite{lee2019dancing,li2022danceformer,li2021ai}, presents a significant challenge across multiple domains, including gaming, filmmaking, and robotic animation. Notably, motion generation based on language descriptions has garnered substantial interest, given its promise for enhancing realism and broadening practical applications.

However, generating a high-quality motion is not trivial due to the inherent modality gap and the complex mapping between text and motion modalities. Previous works \cite{petrovich2022temos, tevet2022motionclip} align the latent feature space between text and motion modalities. For instance, TEMOS~\cite{petrovich2022temos} aligns the text and motion feature by learning text and motion encoders. MotionClip~\cite{tevet2022motionclip} aligns the human motion manifold to CLIP~\cite{radford2021learning} space for infusing semantic knowledge of CLIP into the motion extractor. Despite these advancements, they might encounter performance degradation when dealing with complex textual descriptions.

To process these complex textual descriptions, some text-to-motion methods are proposed. State-of-the-art approaches such as TM2T~\cite{guo2022tm2t} and T2M-GPT~\cite{zhang2023t2m} use vector-quantized autoencoder (VQ-VAE) \cite{van2017neural} to learn a discrete and compact motion representation, followed by a translation model \cite{radford2018improving, vaswani2017attention} to map the text modality to the motion modality. With the popularity and the superior performance in the generation tasks of diffusion models \cite{ho2020denoising}, MDM~\cite{tevet2022human} and MLD~\cite{chen2022executing} are proposed to learn conditioned diffusion models on the raw motion representation space and the latent feature space, respectively. 

While there have been promising advancements, two primary issues remain unresolved: i) The aforementioned diffusion methods, namely MDM~\cite{tevet2022human} and MLD~\cite{chen2022executing}, predominantly address the latent feature within a continuous space. Although VQ-VAE-inspired architectures have made considerable strides in motion generation \cite{zhang2023t2m, guo2022tm2t}, particularly with the support of discrete and compact motion representations, the integration of the diffusion model into a discrete space remains inadequately explored. ii) Discrete diffusion models employed in prior studies \cite{gu2022vector,gong2022diffuseq,hoogeboom2021argmax} treat all tokens uniformly. This approach presupposes that every token conveys an equivalent amount of information, neglecting the inherent disparities among tokens within a sequence. A more intuitive generative approach for humans would involve a progressive hierarchy, commencing with overarching concepts and gradually delving into finer details.

To address the aforementioned challenges, we introduce a priority-centric motion discrete diffusion model (M2DM) designed to generate motion sequences from textual descriptions, progressing in a primary to secondary manner. Initially, we employ a Transformer-based VQ-VAE, which is adept at learning a concise, discrete motion representation through the global self-attention mechanism. To circumvent code collapse and guarantee the optimal utilization of each motion token within the codebook, a regularization term is incorporated during the VQ-VAE training phase. Furthermore, we craft a noise schedule wherein individual tokens within a sequence are allocated varying corruption probabilities, contingent on their respective priorities during the forward process. Specifically, tokens of higher priority are slated for corruption towards the latter stages of the forward process. This ensures that the subsequent learnable reverse process adheres to a primary-to-secondary sequence, with top-priority tokens being reinstated foremost.

To discern the significance of motion tokens within a sequence, we introduce two evaluative strategies: static assessment and dynamic assessment. 1) Static assessment: Drawing inspiration from the neural language domain, where the significance of individual words is gauged by their entropy across datasets, we calculate the entropy of each motion token across the entire motion dataset. 2) Dynamic assessment: We cultivate an agent specifically to dynamically gauge the importance of tokens within a sequence. Given a discrete motion token sequence, the agent masks one token at each iteration. These masked token sequences are then fed into the VQ decoder for continuous motion reconstruction. The agent's objective is to curtail the cumulative reconstruction discrepancy between the original and the reconstructed motions at every phase. This is achieved using a reinforcement learning~\cite{kaelbling1996reinforcement} (RL) strategy, facilitating the agent's identification of motion tokens that convey minimal information within a sequence—a process we term dynamic analysis.
Our priority-centric discrete diffusion model has showcased commendable generative prowess, especially when dealing with intricate textual descriptions. Through rigorous testing, our method has proven its mettle, consistently matching or surpassing the performance of prevailing text-to-motion generation techniques on the HumanML3D and KIT-ML datasets.

We summarize our contributions as follows:
\begin{itemize}
    \item To capture long-range dependencies in the motion sequences, we apply Transformer as the architecture of VQ-VAE. Besides, a regularization term is applied to increase the usage of tokens in the codebook.
    \item We design a priority-centric scheme for human motion generation in the discrete latent space. To enhance the performance given complex descriptions, we design a novel priority-centric scheme for the discrete diffusion model. 
    \item Our proposed priority-centric M2DM achieves state-of-the-art performance on the HumanML3D~\cite{guo2022generating} and KIT-ML~\cite{plappert2016kit} datasets.
\end{itemize}

\section{Related Work}

\paragraph{VQ-VAE.}
Vector Quantized Variational Autoencoder (VQ-VAE) \cite{van2017neural} aims to minimize reconstruction error with discrete representations by training a variational autoencoder. VQ-VAE achieves promising performance on generative tasks, such as text-to-image generation \cite{ramesh2021zero, yu2022scaling}, music generation \cite{dhariwal2020jukebox}, unconditional image synthesis \cite{esser2021taming}, etc. For motion generation tasks, most of the methods \cite{guo2022tm2t, zhang2023t2m} tokenize the continuous motion representation by the 1D-CNN network. Since the motion style varies from person to person and the number of action combinations is almost infinite, it is essential to extract discriminative and distinctive motion features from the raw motion representations. Moreover, raw motion representation is much more redundant compared with neural language, it is feasible to represent a motion sequence using few motion tokens from a global perspective. To this end, we introduce a transformer-based VQ-VAE and several techniques are applied to balance the usage of each motion token in the codebook and prevent the codebook from collapsing, \textit{i.e.}, only a small proportion of codes in the codebook are activated.

\paragraph{Diffusion models.}

Diffusion generative models are first proposed in \cite{sohl2015deep} and achieve promising results on image generation \cite{dhariwal2021diffusion, nichol2021improved, ho2022cascaded,ho2020denoising,XingyiCVPR23,XingyiICCV23,RunpengCVPR23,fang2023structural,fang2023depgraph}. Discrete diffusion model is first proposed by \cite{sohl2015deep} and several works \cite{hoogeboom2021argmax, gong2022diffuseq} apply the discrete diffusion model for text generation. D3PM \cite{austin2021structured} then applies the discrete diffusion model to the image generation task. For motion generation tasks, several works \cite{kim2022flame, zhang2022motiondiffuse, tevet2022human, chen2022executing} introduce diffusion models to generate diverse motion sequences from textual descriptions. In this work, we first introduce the discrete diffusion model for text-to-motion generation tasks which achieves promising results, especially for long text descriptions with several action commands.

\paragraph{Conditional human motion synthesis} 

Compared with unconditional Human motion synthesis, human motion synthesis conditioned on various conditions, such as textual description~\cite{zhang2023t2m, guo2022tm2t, chen2022executing, zhang2022motiondiffuse}, music~\cite{lee2019dancing,li2022danceformer,li2021ai,Gong2023TM2D}, and actions~\cite{petrovich2021action, guo2020action2motion} is more challenging due to the huge gap between two different domains. A common strategy of conditioned human motion synthesis is to employ generative models, for instance, conditional VAE \cite{kingma2013auto}, and learn a latent motion representation. Text-to-motion task is more challenging because neural language is a type of information highly refined by human beings, whereas human motion is much more redundant. Most recently, several promising works \cite{chen2022executing, guo2022generating, guo2022tm2t, zhang2022motiondiffuse, zhang2023t2m} are proposed for text-to-motion tasks. T2M-GPT \cite{zhang2023t2m} and TM2T \cite{guo2022tm2t} regard the text-to-motion task as a translation problem. They first train a VQ-VAE \cite{van2017neural} to tokenize the motion sequences into motion tokens, and Recurrent Neural Networks (RNN) \cite{medsker2001recurrent}, Transformer \cite{vaswani2017attention,yang2022dery,DBLP:conf/eccv/YangYW22,SuchengShunted22,SGFormerICCV23,MetaFormerBasline}, or GPT-like \cite{radford2018improving} models are further applied to `translate' textual descriptions to motions. Moreover, diffusion-based \cite{ho2020denoising} models are introduced for text-to-motion generation by \cite{zhang2022motiondiffuse, dabral2022mofusion, tevet2022human, chen2022executing} in the contentious space. In this work, to the best of our knowledge, we are the first ones to diffuse the motion representation in the discrete latent space. Moreover, we notice that different motion clips in a whole motion sequence play various roles, for instance, stepping forward is more important than swinging arms for a running motion. We further design a diffuse schedule such that the generation process follows a from-primary-to-secondary manner.

\section{Method}

\begin{figure*}[!htbp] 
\centering 
\includegraphics[width=0.96\textwidth]{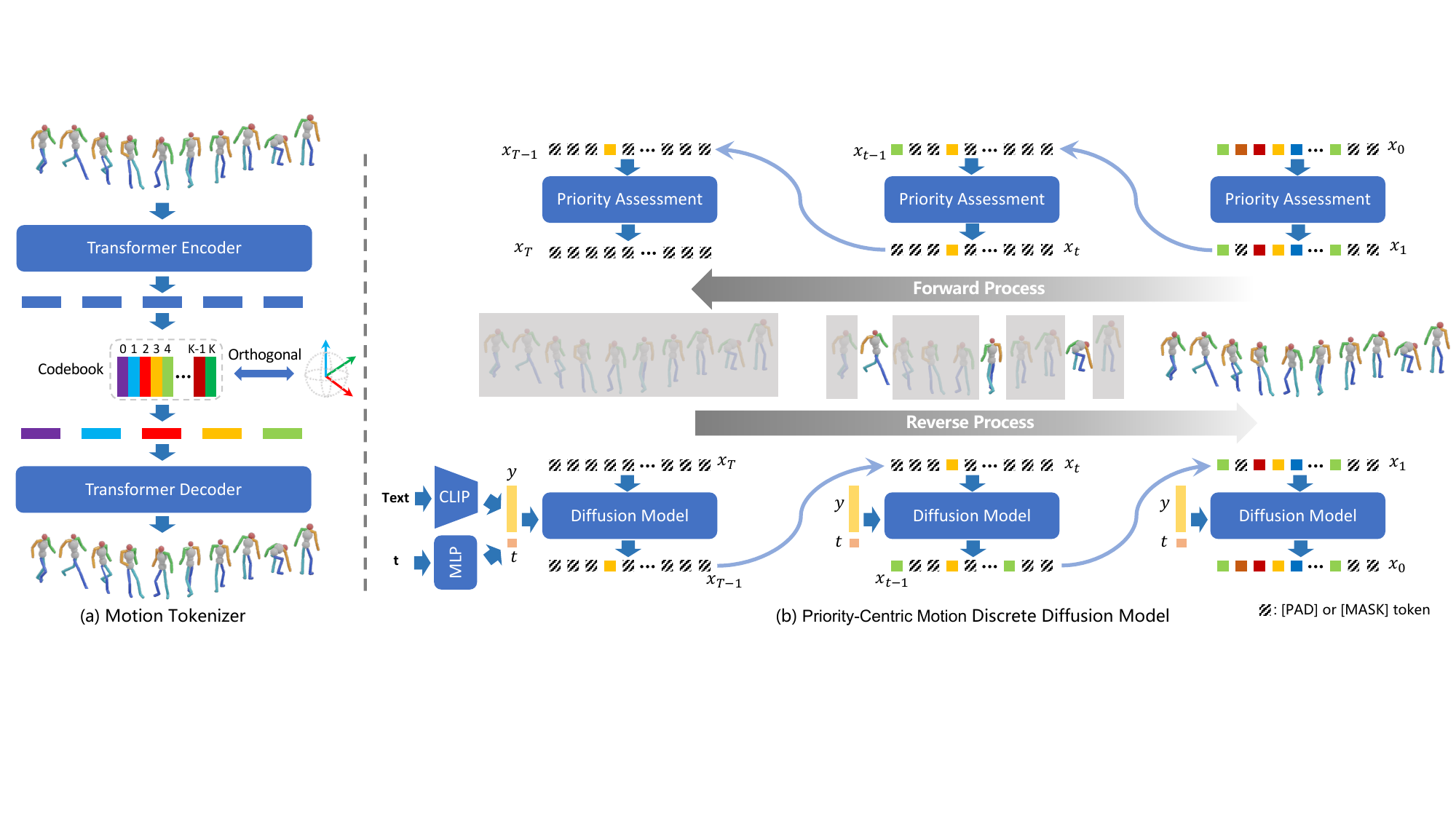} 
\caption{\textbf{Overview of our framework for our Motion Discrete Diffusion Model (M2DM).} M2DM consists of a Transformer-based VQ-VAE (Sec.~\ref{vqvae}) and a discrete diffusion model (Sec. \ref{diffusion}). Firstly, the Transformer-based motion tokenizer is learned to discretize motion sequences into tokens and reconstruct original motion sequences from tokens queried from the learnable codebook. Then the discrete diffusion model is learned to generate diverse motions conditioned on the textual descriptions. Two priority assessment strategies (Sec. \ref{denoising_process}) are applied to mask or replace motion tokens with less information at the beginning of the forward diffusion process and the most important motion tokens are corrupted in the last few diffusion steps. The diffusion model learns to restore the original motion token index conditioned on the given textual descriptions. }
\label{overall} 
\end{figure*}

\subsection{Motion Tokenizer}
\label{vqvae}

The motion tokenizer aims to learn a codebook that can represent a wide range of motion sequences with the priority awareness of motion tokens. To achieve this, we propose a transformer-based VQ-VAE that leverages self-attention to capture long-range dependencies between motion frames and learns a stable and diverse codebook. 

Fig.~\ref{overall} shows the overview of our motion tokenizer.
Given a motion sequence $X\in \mathbb{R}^{T \times d}$ with $T$ frames and $d$ dimensional motion representation, we reconstruct the motion sequence using a transformer autoencoder and learn an informative codebook  $Z=\left \{ z_k \right \} _{k=1}^K$ with $K$ entries. 
We use the transformer encoder $E$ to compute the latent motion features $\boldsymbol{b}=\left \{ b_t \right \}_{t=1}^{T} \in \mathbb{R}^{T\times d'}$, where $d'$ is the dimension of the latent features. Then for each feature $b_t$, we query the codebook $Z$ by measuring the distance between feature $b_t$ and each entry in the codebook $Z$. To improve the codebook usage and avoid dead codes, we first normalize $b_t$ and $z_t$ and then measure the Euclidean distance between two normalized vectors in the unit sphere coordinate, which is formulated as:

\begin{equation}
    \hat{b} _t= \underset{b_t\in \boldsymbol{b}}\arg \min \left \| \ell _2(b_t) - \ell_2(z_k) \right \|_2^2 ,
\end{equation}
where $\ell_2$ is the $l_2$ normalized function.

The standard optimization goal consists of three parts: the reconstruction loss, the embedding loss, and the commitment loss, which is formulated as
\begin{equation}
    \begin{aligned}
    \mathcal{L}_{vq} &= \left \| X - \tilde X \right \|_2^2  + ||Z - \mathit{sg}[\hat{Z}]||_2^2 \\
    &+ \eta  ||\mathit{sg}[Z] - \hat{Z}||_2^2.
    \end{aligned}
\end{equation}

Moreover, to achieve an informative codebook with diverse and high usage of its entries, we apply an orthogonal regularization. Specifically, we calculate the distance between each pair of codebook entries to enforce orthogonality among all the codebook entries:
\begin{equation}
    \mathcal{L}_{orth}(Z) =  \left \| \ell_2(Z)^\top \ell_2(Z) - I_K  \right \| _2^2,
\end{equation}
where $\mathcal{L}_2$ is the $l_2$ normalized function.

Therefore, the overall training objective of vector quantization is defined as follows:
\begin{equation}
    \begin{aligned}
    \mathcal{L}_{vq} &= \left \| X - \tilde X \right \|_2^2  + ||Z - \mathit{sg}[\hat{Z}]||_2^2 \\
    &+ \eta  ||\mathit{sg}[Z] - \hat{Z}||_2^2 +  \delta \mathcal{L}_{orth}(Z),
    \end{aligned}
\end{equation}
where $\tilde X = D(\boldsymbol{b})$, $D$ is the decoder, $\mathit{sg}[\cdot]$ denotes the stop-gradient operation, $\eta $ and $\delta$ are the weighting scalars.

\subsection{Discrete Diffusion Models}
\label{diffusion}

Generally speaking, the forward discrete diffusion process corrupts each element $x_t$ at timestep $t$ with $K$ categories via fixed Markov chain $q(x_t | x_{t-1})$, for instance, replace some tokens of $x_{t-1}$ or mask the token directly.
After a fixed number of $T$ timesteps, the forward process yields a sequence of increasingly noisy discrete tokens $z_1, ..., z_T$ of the same dimension of $z_0$ and $z_T$ becomes pure noise token.
For the denoising procedure, the real $x_0$ is restored sequentially based on the pure noisy token $z_T$.
To be specific, for scalar random variable with $K$ categories $x_t, x_{t-1} \in 1, ... K$, the forward transition probabilities can be represented by the transition matrices $[\boldsymbol{Q}_t]_{ij}=q(x_t=i | x_{t-1}=j) \in \mathbb{R}^{(K+1) \times (K+1)}$. $[\boldsymbol{Q}_t]_{i,j}$ is the Markov transition matrix from state $t$ to state $t-1$ that is applied to each token in the sequence independently, which can be written as:
\begin{equation}
\label{Q_equation}
    \boldsymbol{Q}_t = \begin{bmatrix}
 \alpha_t+\beta_t & \beta_t & \beta_t & \dots  & 0\\
 \beta_t &  \alpha_t+\beta_t & \beta_t & \dots & 0\\
 \beta_t & \beta_t &  \alpha_t+\beta_t & \dots & 0\\
 \vdots  & \vdots  & \vdots  &  & \vdots  \\
 \gamma_t & \gamma_t & \gamma_t & \dots &1
\end{bmatrix},
\end{equation}
where $\alpha_t \in [0,1]$ is the probability of retaining the token, $\gamma_t$ is the probability of replacing the original token to $\mathrm{[MASK]}$ token, leaving the probability $\beta_t=(1-\alpha_t-\gamma_t)/K$ to be diffused. The forward Markov diffusion process for the whole toke sequence is written as,
\begin{equation}
    q(\boldsymbol{x}_t|\boldsymbol{x}_{t-1}) = \mathrm{Cat}(\boldsymbol{x};\boldsymbol{p}=\boldsymbol{x}_{t-1}\boldsymbol{Q}_t) = \boldsymbol{x}_t\boldsymbol{Q}_t\boldsymbol{x}_{t-1},
\end{equation}
where $\boldsymbol{x}$ is the one-hot row vector identifying the token index and $\mathrm{Cat}(\boldsymbol{x};\boldsymbol{p})$ is a categorical distribution over the one-hot vector $\boldsymbol{x}_t$ with probabilities $\boldsymbol{p}$. 

Starting from the initial step $\boldsymbol{x}_0$, the posterior diffusion process is formulated as:
\begin{equation}
    \begin{aligned}
    q(\boldsymbol{x}_{t-1}|\boldsymbol{x}_t,\boldsymbol{x}_0)
    &=\frac{q(\boldsymbol{x}_t|\boldsymbol{x}_{t-1}, \boldsymbol{x}_0)q(\boldsymbol{x}_{t-1},\boldsymbol{x}_0)}{q(\boldsymbol{x}_t|\boldsymbol{x}_0)} \\
    &=\frac{(\boldsymbol{x}^\top \boldsymbol{Q}_t\boldsymbol{x}_{t-1})(\boldsymbol{x}_{t-1}^{T}\bar{\boldsymbol{Q}} _{t-1}\boldsymbol{x}_0)}{\boldsymbol{x}_t^\top \bar{\boldsymbol{Q}} _t\boldsymbol{x}_0}
    \end{aligned}
\end{equation}
with $\bar{\boldsymbol{Q}}=\boldsymbol{Q}_1\boldsymbol{Q}_2\dots \boldsymbol{Q}_t$.

Recording to Markov chain, the intermediate step can be marginalized out and the derivation of the probability of $\boldsymbol{x}_t$ at arbitrary timestep directly from $\boldsymbol{x}_0$ can be formulated by
\begin{equation}
    q(\boldsymbol{x}_t|\boldsymbol{x}_0)=\boldsymbol{x}_t^\top\bar{\boldsymbol{Q}}\boldsymbol{x}_0.
\end{equation}

The cumulative transition matrix $\bar{\boldsymbol{Q}}$ and the probability $q(\boldsymbol{x}_t|\boldsymbol{x}_0)$ are then computed as
\begin{equation}
    \bar{\boldsymbol{Q}}\boldsymbol{x}_0=\bar\alpha_t\boldsymbol{x}_0+(\bar\gamma - \bar\beta_t)\boldsymbol{m}+\bar\beta_t
\end{equation}
where $\boldsymbol{m}$ is the one-hot vector for $\mathrm{[MASK]}$ and  $\mathrm{[PAD]}$, $\bar\alpha_t=\prod_{i=1}^{t}\alpha_i $, $\bar\gamma_t = 1-\prod_{i=1}^{t} (1-\gamma_i)$, and $\bar\beta_t=(1-\bar\alpha_t-\bar\gamma_t)/K$.

Regarding the reverse diffusion process, we train a transformer-like \cite{vaswani2017attention} denoising network $p_{\theta}(\boldsymbol{x}_{t-1}|\boldsymbol{x}_t,\boldsymbol{y})$ to estimate the posterior transition distribution $q(\boldsymbol{x}_{t-1}|\boldsymbol{x}_t,\boldsymbol{x}_0)$.
The network is trained to minimize the variational lower bound (VLB) \cite{sohl2015deep}:
\begin{equation}
\begin{aligned}
   \mathcal{L} _{VLB}=&\mathbb{E}_q[D_{KL}(q(\boldsymbol{x}_T|\boldsymbol{x}_0) \parallel p_{\theta}(\boldsymbol{x}_T))] \\
&+\mathbb{E}_q[\sum_{t=2}^{T} D_{KL}(q(\boldsymbol{x}_{t-1}|\boldsymbol{x}_t,\boldsymbol{x}_0) \parallel  p_{\theta}(\boldsymbol{x}_{t-1}|\boldsymbol{x}_t,t))] \\
&-\log p_{\theta}(\boldsymbol{x}_0|\boldsymbol{x}_1)
\end{aligned}
\end{equation}
where $\mathbb{E}_q[\cdot]$ denotes the expectation over the joint distribution $q(\boldsymbol{x}_{0:T})$.

\subsection{Priority-Centric Denoising Process}
\label{denoising_process}

The designation of the noise schedule in the continuous domain, for instance, the linear schedule \cite{ho2020denoising} and the cosine schedule \cite{nichol2021improved}, achieves the excellent performance of diffusion models. The noise can be easily controlled by the variation of Gaussian noise. For the discrete diffusion models, several noise schedules \cite{hoogeboom2021argmax, austin2021structured} have been explored to control the data corruption and denoise the reverse process by choosing the transition matrix $\boldsymbol{Q}_t$. \textit{For text-to-motion generation, however, such a schedule assumes all motion tokens carry the same amount of information and do not consider the difference among the discrete tokens extracted from the continuous motion sequence.} Intuitively, the motion generation process would be in a progressive manner, where the most important motion tokens should appear in earlier steps for laying a solid foundation when denoising the noise in the discrete domain. 

Specifically, we apply the mask-and-replace strategy in Eq.~\eqref{Q_equation} to corrupt each ordinary discrete token with a probability of $\gamma_t$ for masking $[\mathrm{MASK}]$ token and a probability of $\beta_t$ for uniform diffusion to a random token $\boldsymbol{x}_k$, resulting in a remaining probability of $\alpha_t=1-K\beta_t-\gamma_t$ for retention. 
When corrupting across the forward process in the discrete domain, we expect that $\bar\alpha_t^i < \bar\alpha_t^j$ if token $\boldsymbol{x}_i$ is more informative than $\boldsymbol{x}_j$ such that the tokens with high information emerge earlier than less informative motion tokens in the reverse process.
Instead of using a normal corruption strategy with linearly increasing $\bar\gamma_t$ and $\bar\beta_t$, we propose a priority-score function $F$ to assess the relative importance of information for a motion token sequence, which enables the recovery of the most informative tokens during the reverse process.

Given an original motion token sequences $\boldsymbol{x}_0$ at timestep 0 with length $N$, we aim to analyze the importance of each token $x_0^i$ in $\boldsymbol{x}_0$. Based on the information entropy theory, we calculate the information entropy for each motion token $x_0^i$ as the importance score
\begin{equation}
    F(x_0^i) = \frac{NH(x_0^i)}{\sum_{j=1}^{N} H(x_0^j)},
\end{equation}
where
\begin{equation}
    H(x)=-\sum_{i=1}^{K} p(x_i)\log p(x_i),
\end{equation}
and $K$ denotes the number of categories belonging to $x_i$. 

After obtaining the importance score function and setting linearly increasing schedule $\bar\gamma_t^i$ and $\bar\beta_t^i$ for all tokens, $\bar\gamma_t^i$ and $\bar\beta_t^i$ for specific token $x_0^i$ can be further updated by
\begin{equation}
    \bar\gamma_t^i \Leftarrow  \bar\gamma_t^i \cdot \sin{\frac{t\pi}{T}} \cdot F(x_0^i),
    \label{sigma}
\end{equation}
and 
\begin{equation}
    \bar\beta^i \Leftarrow  \bar\beta^i \cdot \sin{\frac{t\pi}{T}} \cdot F(x_0^i),
    \label{beta}
\end{equation}
where $\boldsymbol{x}_0^i$ is the $i$-th token at timestep 0.

Now we introduce two solutions for obtaining $p(x_i)$.

\paragraph{Static assessment.}

Analogously to computing the entropy for each word in natural language processing, we estimate the probability $p(x_i)$ for each motion token by counting its frequency in the quantized motion sequences over the entire dataset.

\begin{figure}[htbp] 
\centering 
\includegraphics[width=0.32\textwidth]{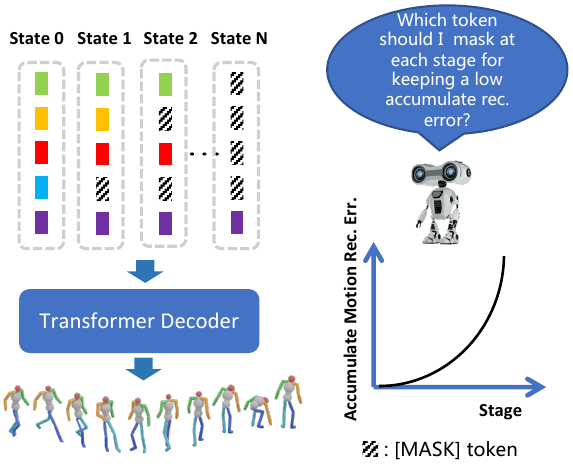} 
\caption{Illustration for dynamic assessment strategy. The gray blocks are masked tokens and other colorful blocks are motion tokens.} 
\label{agent} 
\end{figure}

\paragraph{Dynamic assessment.}
Rather than counting motion token frequency and calculating the importance of each token, we further propose to learn an agent to estimate the importance of each token in the token sequences by reinforcement learning strategy. As shown in Fig.~\ref{agent}, given a series of complete motion tokens $\boldsymbol{x}=\left \{ x_n \right \} _{n=1}^N$ that are queried from the codebook at current state $\boldsymbol{s}_t \in \mathcal{S}$, the scorer samples the motion tokens $\boldsymbol{\tilde x}=\left \{ x_n \right \} _{n=1}^{N'} $ and tries to minimize the reconstruction error between the original motion sequences $\boldsymbol{m}$ and the motion sequences decoded from the sampled tokens $\boldsymbol{\tilde m}$. 

State includes the currently selected motion tokens $\boldsymbol{\tilde x}=\left \{ x_n \right \} _{n=1}^{N'} $, the complete motion tokens $\boldsymbol{x}=\left \{ x_n \right \} _{n=1}^N$, and the corresponding reconstructed continuous motions $\boldsymbol{\tilde m}$ and $\boldsymbol{m}$. The reconstructed continuous motion sequences $\boldsymbol{\tilde m}$ and $\boldsymbol{m}$ are obtained by decoding the motion tokens by VQ decoder. The action is to sample one motion token from the complete motion token sequence at each stage. To be specific, given a motion token sequence, the agent $\pi_{\theta}(\boldsymbol{a}_t|\boldsymbol{s}_t)$ tries to sample a motion token that can recover the continuous motion sequences with the minimum reconstruction error. The reward measures the reconstruction difference between $\boldsymbol{\tilde m}$ and $\boldsymbol{m}$ with the minimum sample times.

\begin{figure}[htbp] 
\centering 
\includegraphics[width=0.45\textwidth]{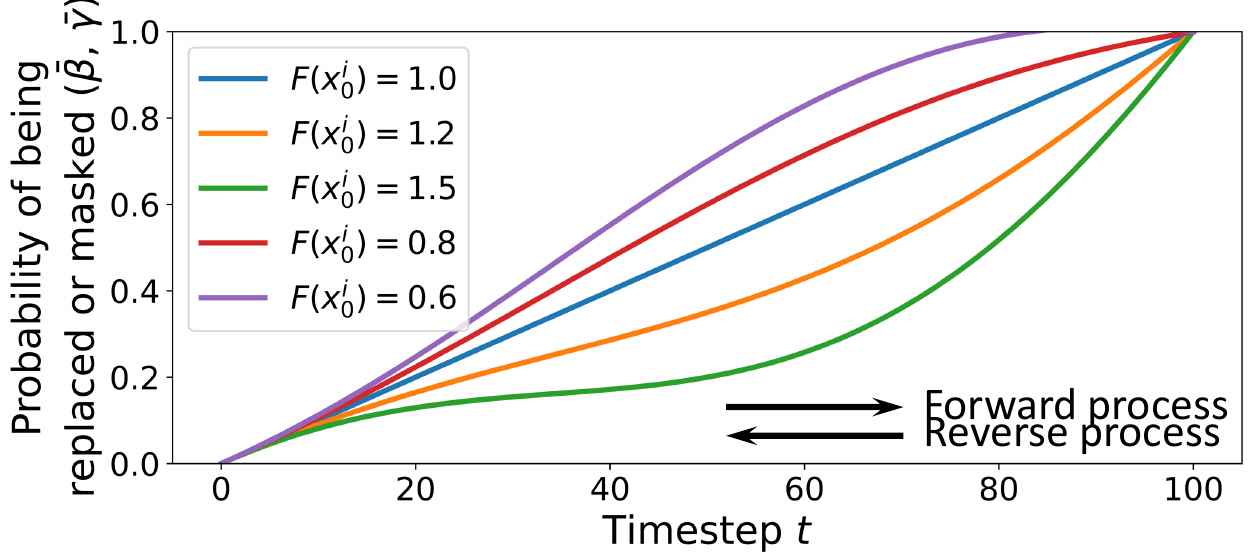} 
\caption{Priority-centric denoising schedule for motion tokens.} 
\label{schedule} 
\end{figure}

After the agent is well-trained, we calculate and restore the sampling order for each motion token sequence by the agent $\pi_{\theta}(\boldsymbol{a}_t|\boldsymbol{s}_t)$ and compute the noise schedule for each motion token in the codebook by Eq.~\eqref{sigma} and \eqref{beta}. Fig. \ref{schedule} shows the noise schedule for different motion tokens \textit{w.r.t.} the corresponding priorities. As shown in Fig.~\ref{schedule}, high-priority tokens (orange and green curves) are encouraged to be corrupted at the end of the forward process such that the learnable reverse process follows a primary-to-secondary manner and high-priority tokens will be restored first. It is worth noting that both strategies are applied to design noise schedules for each token, which can be calculated and stored in advance such that the agent will not join the forward and reverse diffusion processes. Thus, there is no extra computational cost for the diffusion model during the training and inference stages.

\section{Experiments}

\subsection{Datasets and Evaluation Metrics}
\label{dataset}

Our experiments are conducted on two standard text-to-motion datasets for text-to-motion generation: HumanML3D~\cite{guo2022generating} and KIT-ML~\cite{plappert2016kit}. We follow the evaluation metrics provided in \cite{guo2022generating}. 
\paragraph{HumanML3D.}

HumanML3D \cite{guo2022generating} is currently the largest 3D human motion dataset that covers a broad range of daily human actions, for instance, exercising and dancing. The dataset contains 14,616 motion sequences and 44970 text descriptions. 
The motion sequences are processed to 20 frame-per-second (FPS) and cropped to 10 seconds if the motion sequences are longer than 10 seconds, resulting in duration ranges from 2 to 10 seconds. 
For each motion clip, the corresponding number of descriptions is at least three. 

\paragraph{KIT Motion-Language (KIT-ML).}
KIT-ML dataset \cite{plappert2016kit} contains 3,911 3D human motion clips with 6,278 text descriptions. For each motion clip, one to four textual descriptions are provided. 
The motion sequences are collected from the KIT dataset \cite{mandery2015kit} and the CMU dataset \cite{cmu} with the down-sampled 12.5 FPS. 
 
Both of the aforementioned datasets are split into training, validation, and testing datasets with the proportions of 80\%, 5\%, and 15\%. 

\paragraph{Implementation details.}

Both our VQ-encoder and decoder consist of 4 transformer layers with 8 heads and the dimension is 512. For the HumanML3D~\cite{guo2022generating} and KIT-ML~\cite{plappert2016kit} datasets, the motion sequences are cropped to 64 frames for training. We apply Adam with $\beta_1=0.9$, $\beta_2=0.999$, weight decay is $1$e-$4$. The learning rate is $1$e-$4$ after warmup linearly for 5 epochs. The number of tokens in the codebook is 8192. We train our VQ-VAE with a batch size of 1024 across 8 Tesla T4 for a total 100 training epochs for 8 hours.

For the discrete diffusion model, we set timesteps $T=100$ and the network is trained using Adam \cite{kingma2014adam} with $\beta_1=0.9$, $\beta_2=0.999$. The learning rate reaches $2$e-$4$ after $2000$ iterations of a linear warmup schedule. The transformer-based discrete diffusion model consists of 12 transformer layers with 8 heads, and the dimension is 512. The text features are extracted by pre-trained CLIP~\cite{radford2021learning} model. The diffusion model is trained with a batch size of 64 across 8 Tesla T4 for 100k iterations for 36 hours.

\paragraph{Evaluation metrics.}

Following the evaluation protocols provided in previous works \cite{guo2022generating, zhang2023t2m, tevet2022human, chen2022executing}, we evaluate the correspondences between motion and language using deep multimodal features with the pre-trained models in \cite{guo2022generating} by the following metrics: 1) Generation diversity: We randomly sample 300 pairs of motions from the motion pool and extracted the selected motion features by the extractor provided by~\cite{guo2022generating}. Euclidean distances of the motion pairs are computed for measuring the motion diversity. 2) R-Precision: Provided one motion representation and 32 textual descriptions, the Euclidean distances between the description feature and each motion feature in the candidates are calculated and ranked. Top-k $(k=1,2,3)$ average accuracy of motion-to-text retrieval is reported. 3) Frechet Inception Distance (FID) \cite{heusel2017gans}: FID is the principal metric to evaluate the feature descriptions between the generated motions and the ground-truth motions by the feature extractor~\cite{guo2022generating}. 4) Multimodal Distance (MM-Dist): MM-Dist measures the average Euclidean distances between each text feature and the generated motion feature. 5) MModality measures the generation diversity within the same textual descriptions. Please refer to the supplementary material for a detailed introduction.

\subsection{Comparisons with state-of-the-art approaches}

We compare our methods with other state-of-the-art methods~\cite{lin:vigil18, ahuja2019language2pose, bhattacharya2021text2gestures, ghosh2021synthesis, tulyakov2018mocogan, lee2019dancing, guo2022tm2t, guo2022generating, tevet2022human, zhang2022motiondiffuse, zhang2023t2m, chen2022executing} on the HumanML3D~\cite{guo2022generating} and KIT-ML~\cite{plappert2016kit} datasets.

\begin{table*}[htbp]
    \centering
    \scalebox{0.85}{

    \begin{tabular}{l c c c c c c c}
    \toprule
    \multirow{2}{*}{Methods}  & \multicolumn{3}{c}{R-Precision $\uparrow$} & \multirow{2}{*}{FID $\downarrow$} & \multirow{2}{*}{MM-Dist $\downarrow$} & \multirow{2}{*}{Diversity $\to$} & \multirow{2}{*}{MModality $\uparrow$}\\

    \cline{2-4}
    ~ & Top-1 & Top-2 & Top-3 \\

    \midrule

        \textbf{Real motion} & \et{0.511}{.003} & \et{0.703}{.003} & \et{0.797}{.002} & \et{0.002}{.000} & \et{2.974}{.008} & \et{9.503}{.065} & -  \\
        \small{Our VQ-VAE} & \et{0.508}{.002} & \et{0.691}{.002} & \et{0.791}{.003} & \et{0.063}{.001}& \et{3.015}{.010} & \et{9.577}{.081} & -  \\
        \small{Our VQ-VAE \textit{w/o} $\ell_2$} & \et{0.494}{.003} & \et{0.685}{.002} & \et{0.772}{.003} & \et{0.070}{.001}& \et{3.251}{.008} & \et{9.525}{.092} & -  \\
        \small{Our VQ-VAE \textit{w/o}} $\ell_2$, $\mathcal{L}_\mathrm{orth}$ & \et{0.485}{.002} & \et{0.671}{.003} & \et{0.752}{.003} & \et{0.079}{.001}& \et{3.378}{.012} & \et{9.511}{.095} & -  \\
    \midrule
        Seq2Seq~\cite{lin:vigil18} & \et{0.180}{.002} & \et{0.300}{.002} & \et{0.396}{.002} & \et{11.75}{.035} & \et{5.529}{.007} & \et{6.223}{.061}  & -  \\

        Language2Pose~\cite{ahuja2019language2pose} & \et{0.246}{.002} & \et{0.387}{.002} & \et{0.486}{.002} & \et{11.02}{.046} & \et{5.296}{.008} & \et{7.676}{.058} & -  \\

        Text2Gesture~\cite{bhattacharya2021text2gestures} & \et{0.165}{.001} & \et{0.267}{.002} & \et{0.345}{.002} & \et{5.012}{.030} & \et{6.030}{.008} & \et{6.409}{.071} & -  \\

        Hier~\cite{ghosh2021synthesis} & \et{0.301}{.002} & \et{0.425}{.002} & \et{0.552}{.004} & \et{6.532}{.024} & \et{5.012}{.018} & \et{8.332}{.042} & -  \\

        MoCoGAN~\cite{tulyakov2018mocogan} & \et{0.037}{.000} & \et{0.072}{.001} & \et{0.106}{.001} & \et{94.41}{.021} & \et{9.643}{.006} & \et{0.462}{.008} & \et{0.019}{.000}  \\

        Dance2Music~\cite{lee2019dancing} & \et{0.033}{.000} & \et{0.065}{.001} & \et{0.097}{.001} & \et{66.98}{.016} & \et{8.116}{.006} & \et{0.725}{.011} & \et{0.043}{.001}  \\

        TM2T~\cite{guo2022tm2t} & \et{0.424}{.003} & \et{0.618}{.003} & \et{0.729}{.002} & \et{1.501}{.017} & \et{3.467}{.011} & \et{8.589}{.076} & \et{2.424}{.093}  \\

        Guo \textit{et al.}~\cite{guo2022generating} & \et{0.455}{.003} & \et{0.636}{.003} & \et{0.736}{.002} & \et{1.087}{.021} & \et{3.347}{.008} & \et{9.175}{.083} & \et{2.219}{.074}  \\

        MDM~\cite{tevet2022human}$^\S$ & - & - & \et{0.611}{.007} & \et{0.544}{.044} & \et{5.566}{.027} & \etr{9.559}{.086} & \etbb{2.799}{.072}  \\

        MotionDiffuse~\cite{zhang2022motiondiffuse}$^\S$ & \et{0.491}{.001} & \etbb{0.681}{.001} & \etr{0.782}{.001} & \et{0.630}{.001} & \etr{3.113}{.001} & \etbb{9.410}{.049} & \et{1.553}{.042}  \\

        T2M-GPT~\cite{zhang2023t2m} & \etbb{0.492}{.003} & \et{0.679}{.002} & \etbb{0.775}{.002} & \etr{0.141}{.005} & \et{3.121}{.009} & \et{9.722}{.082} &  \et{1.831}{.048} \\

        MLD~\cite{chen2022executing}  & \et{0.481}{.003} & \et{0.673}{.003} & \et{0.772}{.002} & \et{0.473}{.013} & \et{3.196}{.010} & \et{9.724}{.082} & \et{2.413}{.079}  \\

    \midrule
        M2DM (Ours) \textit{w}. linear schedule & \et{0.452}{.003} & \et{0.678}{.002} & \et{0.752}{.003} & \et{0.417}{.004} & \et{3.167}{.008} & \et{9.972}{.089} & \etbb{3.562}{.071} \\
        M2DM (Ours) \textit{w}. static assessment  & \etbb{0.492}{.003} & \et{0.671}{.003} & \etbb{0.775}{.003} & \et{0.395}{.005} & \etbb{3.116}{.008} & \et{9.937}{.075} &  \et{3.413}{.035} \\
        M2DM (Ours) \textit{w}. dynamic assessment  & \etr{0.497}{.003} & \etr{0.682}{.002} & \et{0.763}{.003} & \etbb{0.352}{.005} & \et{3.134}{.010} & \et{9.926}{.073} &  \etr{3.587}{.072} \\
    \bottomrule
    \end{tabular}
    }

    \footnotesize{ $^\S$ reports results using ground-truth motion length.} \vspace{-3mm}
    \caption{\textbf{Comparison with the state-of-the-art methods on HumanML3D~\cite{guo2022generating} test set.} The evaluation metrics are evaluated by the motion encoder from Guo \textit{et al.}~\cite{guo2022generating}. The right row $\to $ means the closer to the real motion the better.}
    \label{humanml3d}
\end{table*}

\begin{table*}[htbp]
    \centering
    \scalebox{0.85}{

    \begin{tabular}{l c c c c c c c}
    \toprule
    \multirow{2}{*}{Methods}  & \multicolumn{3}{c}{R-Precision $\uparrow$} & \multirow{2}{*}{FID $\downarrow$} & \multirow{2}{*}{MM-Dist $\downarrow$} & \multirow{2}{*}{Diversity $\to$} & \multirow{2}{*}{MModality $\uparrow$}\\

    \cline{2-4}
    ~ & Top-1 & Top-2 & Top-3 \\

    \midrule
 \textbf{Real motion} & \et{0.424}{.005} & \et{0.649}{.006} & \et{0.779}{.006} & \et{0.031}{.004} & \et{2.788}{.012} & \et{11.08}{.097} & -  \\

Our VQ-VAE (Recons.) & \et{0.417}{.004} & \et{0.621}{.003} & \et{0.741}{.006} & \et{0.413}{.009} & \et{2.772}{.018} & \et{10.851}{.105} & -  \\
    \midrule
        Seq2Seq~\cite{lin:vigil18} & \et{0.103}{.003} & \et{0.178}{.005} & \et{0.241}{.006} & \et{24.86}{.348} & \et{7.960}{.031} & \et{6.744}{.106}  & -  \\

        Language2Pose~\cite{ahuja2019language2pose} & \et{0.221}{.005} & \et{0.373}{.004} & \et{0.483}{.005} & \et{6.545}{.072} & \et{5.147}{.030} & \et{9.073}{.100} & -  \\

        Text2Gesture~\cite{bhattacharya2021text2gestures} & \et{0.156}{.004} & \et{0.255}{.004} & \et{0.338}{.005} & \et{12.12}{.183} & \et{6.964}{.029} & \et{9.334}{.079} & -  \\

        Hier~\cite{ghosh2021synthesis} & \et{0.255}{.006} & \et{0.432}{.007} & \et{0.531}{.007} & \et{5.203}{.107} & \et{4.986}{.027} & \et{9.563}{.072} & -  \\

         MoCoGAN~\cite{tulyakov2018mocogan} & \et{0.022}{.002} & \et{0.042}{.003} & \et{0.063}{.003} & \et{82.69}{.242} & \et{10.47}{.012} & \et{3.091}{.043} & \et{0.250}{.009}  \\

        Dance2Music~\cite{lee2019dancing} & \et{0.031}{.002} & \et{0.058}{.002} & \et{0.086}{.003} & \et{115.4}{.240} & \et{10.40}{.016} & \et{0.241}{.004} & \et{0.062}{.002}  \\
        TM2T~\cite{guo2022tm2t} & \et{0.280}{.005} & \et{0.463}{.006} & \et{0.587}{.005} & \et{3.599}{.153} & \et{4.591}{.026} & \et{9.473}{.117} & \et{3.292}{.081}  \\
        Guo \textit{et al.}~\cite{guo2022generating} & \et{0.361}{.006} & \et{0.559}{.007} & \et{0.681}{.007} & \et{3.022}{.107} & \et{3.488}{.028} & \et{10.72}{.145} & \et{2.052}{.107}  \\

        MDM~\cite{tevet2022human}$^\S$ & - & - & \et{0.396}{.004} & \etbb{0.497}{.021} & \et{9.191}{.022} & \et{10.847}{.109} & \et{1.907}{.214}  \\

        MotionDiffuse~\cite{zhang2022motiondiffuse}$^\S$ & \etr{0.417}{.004} & \et{0.621}{.004} & \et{0.739}{.004} & \et{1.954}{.062} & \etr{2.958}{.005} & \etr{11.10}{.143} & \et{0.730}{.013}  \\

        T2M-GPT~\cite{zhang2023t2m} & \etbb{0.416}{.006}  & \et{0.627}{.006} & \etr{0.745}{.006} & \et{0.514}{.029} & \etbb{3.007}{.023} & \etbb{10.921}{.108} & \et{1.570}{.039} \\

        MLD~\cite{chen2022executing}  & \et{0.390}{.003} & \et{0.609}{.003} & \et{0.734}{.002} & \etr{0.404}{.013} & \et{3.204}{.010} & \et{10.8}{.082} & \et{2.192}{.079}  \\

    \midrule

        M2DM (Ours) \textit{w}. linear schedule  & \et{0.405}{.003} & \etr{0.629}{.005} & \et{0.739}{.004} & \et{0.502}{.049} & \et{3.012}{.015} & \et{11.375}{.079} & \et{3.273}{.045}\\
        M2DM (Ours) \textit{w}. static assessment  & \etr{0.417}{.006} & \et{0.625}{.003} & \et{0.741}{.006} & \et{0.521}{.041} & \et{3.024}{.018} & \et{11.373}{.081} & \etbb{3.317}{.031}\\
        M2DM (Ours) \textit{w}. dynamic assessment   & \etbb{0.416}{.004}  & \etbb{0.628}{.004} & \etbb{0.743}{.004} & \et{0.515}{.029} & \et{3.015}{.017} & \et{11.417}{.97} & \etr{3.325}{.37} \\

    \bottomrule
    \end{tabular}
    }
    
    \footnotesize{$^\S$ reports results using ground-truth motion length.} \vspace{-3mm}
    \caption{\textbf{Comparison with the state-of-the-art methods on KIT-ML~\cite{plappert2016kit} test set.} The evaluation metrics are evaluated by the motion encoder from Guo \textit{et al.}~\cite{guo2022generating}. The right row $\to $ means the closer to the real motion the better.}
    \label{kit}
\end{table*}

\begin{table}[htbp]
    \centering
    \scalebox{0.45}{

    \begin{tabular}{c l c c c c c c c}
    \toprule
    \multirow{2}{*}{Text Length}  &  \multirow{2}{*}{Methods}  & \multicolumn{3}{c}{R-Precision $\uparrow$} & \multirow{2}{*}{FID $\downarrow$} & \multirow{2}{*}{MM-Dist $\downarrow$} & \multirow{2}{*}{Diversity $\to$} & \multirow{2}{*}{MModality $\uparrow$}\\

    \cline{3-5}
    ~  & ~ & Top-1 & Top-2 & Top-3 \\

    \midrule

        \multirow{2}{*}{All}  &  TM2T~\cite{guo2022tm2t} & \et{0.424}{.003} & \et{0.618}{.003} & \et{0.729}{.002} & \et{1.501}{.017} & \et{3.467}{.011} & \et{8.589}{.076} & \et{2.424}{.093}  \\
        ~  &  MLD~\cite{chen2022executing}  & \et{0.481}{.003} & \et{0.673}{.003} & \et{0.772}{.002} & \et{0.473}{.013} & \et{3.196}{.010} & \et{9.724}{.082} & \et{2.413}{.079}  \\
        ~  &  M2DM (Ours) & \et{0.497}{.003} & \et{0.682}{.002} & \et{0.763}{.003} & \et{0.352}{.005} & \et{3.134}{.010} & \et{9.926}{.073} &  \et{3.587}{.072} \\

    \midrule

        \multirow{2}{*}{$<$ 15 words}  &  TM2T~\cite{guo2022tm2t} & \et{0.433}{.003} & \et{0.627}{.003} & \et{0.738}{.003} & \et{1.592}{.017} & \et{3.446}{.008} & \et{8.677}{.077} & \et{2.520}{.051}  \\
        ~  &  MLD~\cite{chen2022executing}  & \et{0.492}{.003} & \et{0.680}{.004} & \et{0.779}{.004} & \et{0.469}{.014} & \et{3.191}{.011} & \et{9.770}{.010} & \et{2.586}{.086}  \\
        ~  &  M2DM (Ours)  & \et{0.508}{.003}  & \et{0.685}{.003} & \et{0.767}{.004} & \et{0.347}{.015} & \et{3.129}{.013} & \et{9.915}{.068} & \et{3.594}{.077} \\

    \midrule

        \multirow{2}{*}{15-30 words}  &  TM2T~\cite{guo2022tm2t} & \et{0.409}{.003} & \et{0.596}{.002} & \et{0.708}{.002} & \et{1.425}{.014} & \et{3.528}{.007} & \et{8.000}{.051} & \et{2.477}{.057}  \\
        ~  &  MLD~\cite{chen2022executing}  & \et{0.406}{.005} & \et{0.592}{.005} & \et{0.701}{.006} & \et{0.480}{.020} & \et{3.591}{.020} & \et{9.089}{.083} & \et{2.727}{.094}  \\
        ~  &  M2DM (Ours)  & \et{0.493}{.003}  & \et{0.681}{.004} & \et{0.761}{.004} & \et{0.353}{.031} & \et{3.135}{.014} & \et{9.928}{.073} & \et{3.588}{.081} \\

    \midrule

        \multirow{2}{*}{$>$ 30 words}  &  TM2T~\cite{guo2022tm2t} & \et{0.353}{.009} & \et{0.536}{.009} & \et{0.650}{.007} & \et{1.779}{.061} & \et{3.577}{.024} & \et{7.411}{.071} & \et{2.612}{.055}  \\
        ~  &  MLD~\cite{chen2022executing}  & \et{0.289}{.010} & \et{0.466}{.012} & \et{0.581}{.015} & \et{0.947}{.066} & \et{3.870}{.039} & \et{8.347}{.100} & \et{2.916}{.063}  \\
        ~  &  M2DM (Ours)  & \et{0.491}{.004}  & \et{0.677}{.004} & \et{0.758}{.004} & \et{0.356}{.038} & \et{3.138}{.025} & \et{9.930}{.088} & \et{3.585}{.071} \\

    \bottomrule
    \end{tabular}
    }
    \caption{\textbf{Comparison with the state-of-the-art methods on HumanML3D~\cite{guo2022generating} test set.} The evaluation metrics are evaluated by the motion encoder from Guo \textit{et al.}~\cite{guo2022generating}. The right row $\to $ means the closer to the real motion the better. The first three rows are the evaluation results on the whole testing dataset. The rest are the evaluation results from textual descriptions with different lengths.}
    \label{result}
\end{table}

\paragraph{Quantitative comparisons.}

For each metric, we repeat the evaluation 20 times and report the average with 95\% confidence interval, followed by~\cite{guo2022generating}. Most of the results are borrowed from~\cite{zhang2023t2m}. Tab.~\ref{humanml3d} and Tab.~\ref{kit} summarize the comparison results on the HumanML3D~\cite{guo2022generating} and the KIT-ML~\cite{plappert2016kit} datasets, respectively. Firstly, the metrics scores for our reconstruction motion achieve close performance compared with real motion, which suggests the distinctive and high-quality discrete motion tokens in the codebook learned by our Transformer-based VQ-VAE. The performance of our VQ-VAE degrades when the normalized function $\ell_2$ and the orthogonal regularization term $\mathcal{L}_\mathrm{orth}$ are removed. For text-to-motion generation, our approach achieves compatible performance (R-Precision, FID, and MModality) compared to other state-of-the-art methods. Moreover, compared with MDM \cite{tevet2022human} and MotionDiffuse~\cite{zhang2022motiondiffuse} which evaluate their performance with the ground-truth motion lengths, our model can generate motion sequences with arbitrary lengths conditioned on the given textual descriptions since the $[\mathrm{MASK}]$ and $[\mathrm{PAD}]$ tokens are introduced in the discrete diffusion process.

Furthermore, we conduct more experiments to evaluate the generation performance conditioned on different lengths of textual descriptions. The testing dataset is split into three parts according to the length of the textual descriptions: less than 15 words, between 15 words and 30 words, and more than 30 words. Tab.~\ref{result} shows the quantitative results on the aforementioned three sub-sets. Both TM2T~\cite{guo2022tm2t} and MLD~\cite{chen2022executing} suffer degradation when the textual descriptions are long and complex, especially the descriptions are more than 30 words. Our method significantly outperforms the other two methods with various metrics and there is not much degradation when the descriptions are more complex. The quantitative results demonstrate the effectiveness of our proposed primary-to-secondary diffusion manner.

\begin{figure*}[htbp]
\centering 
\includegraphics[width=1.0\textwidth]{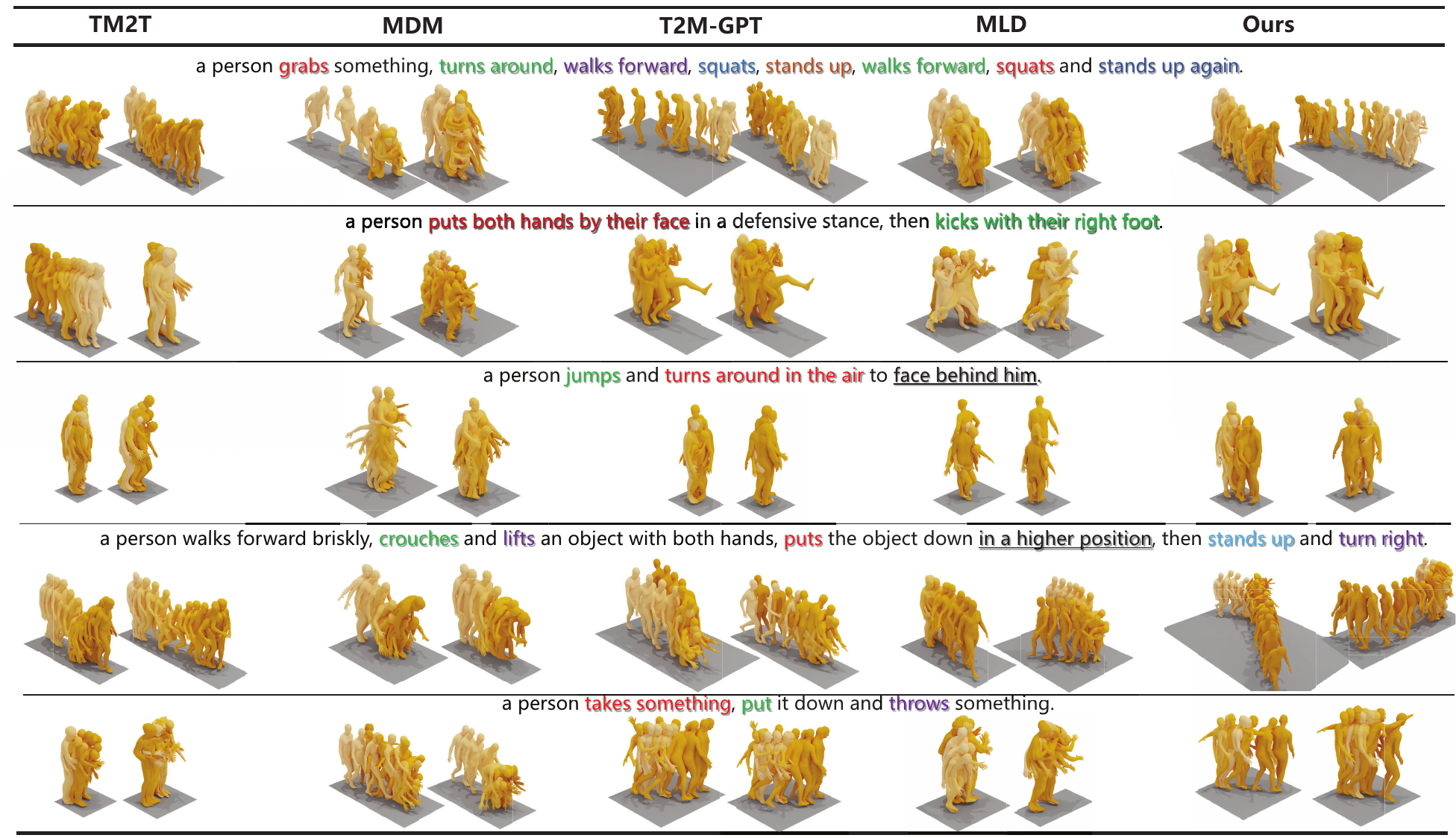} 
\caption{Qualitative comparison of the state-of-the-art methods on the HumanML3D~\cite{guo2022generating} dataset. We compare our generations with TM2T~\cite{guo2022tm2t}, MDM~\cite{tevet2022human}, T2M-GPT~\cite{zhang2023t2m}, and MLD~\cite{chen2022executing}. The color of human mesh goes from light to dark over time.} 
\label{qualitative} 
\end{figure*}

\paragraph{Qualitative comparisons.}

Fig.~\ref{qualitative} qualitatively compares the generated motions from the same textual descriptions. Most of the methods suffer from motion freezing if the textual descriptions consist of several complex commands. For example, the first row in Fig.~\ref{qualitative} shows the generation results from a long textual description with various duplicate action commands. MDM~\cite{tevet2022human} and MLD~\cite{chen2022executing} generate fewer semantic motions compared with the given description. With the help of compact discrete motion tokens, TM2T~\cite{guo2022tm2t} and T2M-GPT~\cite{zhang2023t2m} achieve better results compared with MDM~\cite{tevet2022human} and MLD~\cite{chen2022executing}. Our motion generation results conform to the textual descriptions and exhibit a high degree of diversity compared with other state-of-the-art methods, which validates the efficacy of our proposed method.

\subsection{Ablation studies}

\begin{figure}[htbp]
 \centering
  \begin{subfigure}[b]{0.49\columnwidth}
    \includegraphics[width=\linewidth]{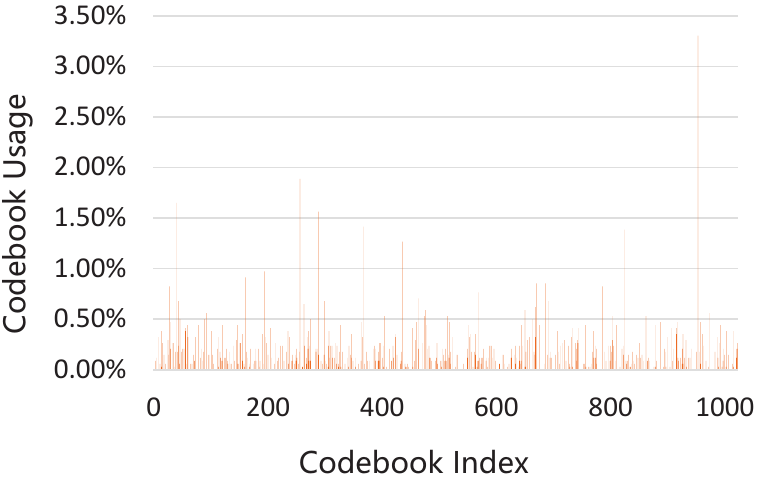}
    \caption{\footnotesize 1D-CNN VQ-VAE.}
    \label{ablation1}
  \end{subfigure}
  \hfill %
  \begin{subfigure}[b]{0.49\columnwidth}
    \includegraphics[width=\linewidth]{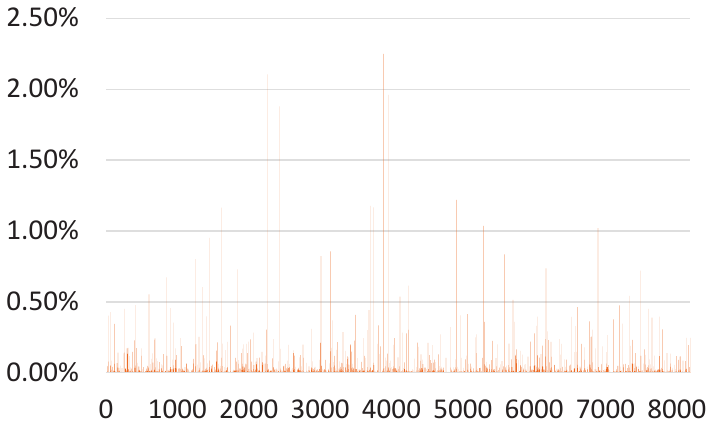}
    \caption{\footnotesize Our VQ-VAE.}
    \label{ablation2}
  \end{subfigure}
  \hfill %
  \begin{subfigure}[b]{0.49\columnwidth}
    \includegraphics[width=\linewidth]{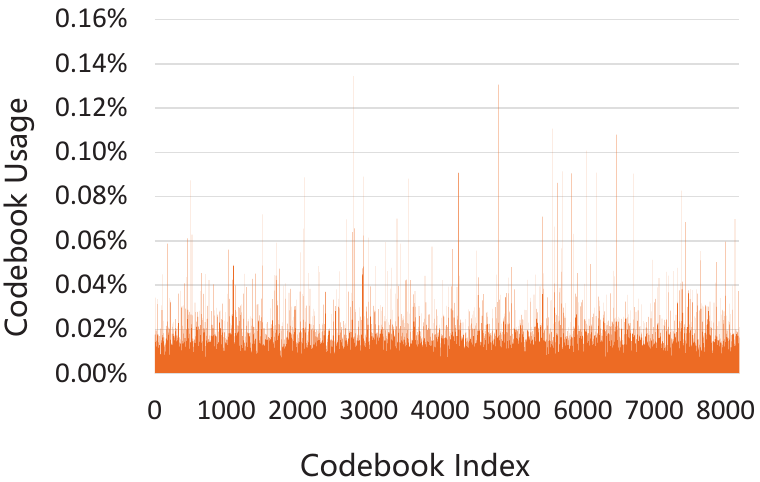}
    \caption{\footnotesize Our VQ-VAE w. $\ell_2$ norm.}
    \label{ablation3}
  \end{subfigure}
  \hfill %
  \begin{subfigure}[b]{0.49\columnwidth}
    \includegraphics[width=\linewidth]{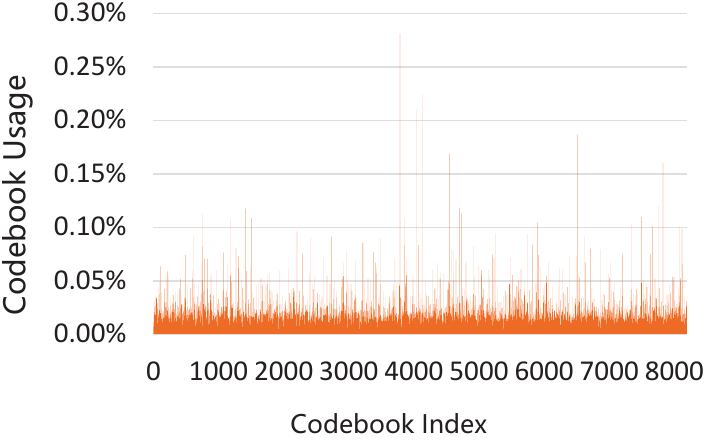}
    \caption{\footnotesize Our VQ-VAE w. $\ell_2$ and $\mathcal{L}_{\mathrm{orth}}$.}
    \label{ablation4}
  \end{subfigure}
  \caption{\textbf{Codebook usage for different training strategies on HumanML3D~\cite{guo2022generating} test set.} We train our VQ-VAE (\textit{Reconstruction}) with different training streategy on the HumanML3D~\cite{guo2022generating} dataset.}
  \label{codebook}
\end{figure}

We delved into the significance of various quantization techniques, as delineated in the initial four rows of Tab.\ref{humanml3d}. The findings indicate that a rudimentary Transformer-based VQ-VAE struggles to reconstruct credible motion sequences, primarily due to the limited efficacy of the codebook. However, the incorporation of the $\ell_2$ norm and $\mathcal{L}_{\mathrm{orth}}$ ensures that each motion token acquires a distinct motion representation. We further scrutinized the utilization of each motion token on the HumanML3D \cite{guo2022generating} test set, with the outcomes depicted in Fig.\ref{codebook}. As evidenced by Fig.\ref{ablation1} and \ref{ablation2}, both the vanilla CNN-based VQ-VAE and Transformer-based VQ-VAE grapple with sub-optimal codebook usage. This is because the tokens in the codebook aren't inherently driven to develop a distinguishing motion representation. Yet, with the aid of the $\ell_2$ norm and $\mathcal{L}_{\mathrm{orth}}$, there's a marked surge in the codebook token utilization. Moreover, when juxtaposing the codebook usage in Fig.\ref{ablation3} with that in Fig.\ref{ablation4}, the latter exhibits a more equitably distributed frequency, attributable to the orthogonal regularization term $\mathcal{L}_{\mathrm{orth}}$.

\section{Conclusion}

In this study, we introduced the priority-centric motion discrete diffusion model (M2DM) tailored for text-to-motion generation. Our model excels in producing varied and lifelike human motions, aligning seamlessly with the input text. This is achieved by mastering a discrete motion representation through a Transformer-based VQ-VAE and implementing a priority-informed noise schedule within the motion discrete diffusion framework. Additionally, we unveiled two innovative techniques for gauging the priority or significance of each motion token. Experimental evaluations across two datasets underscore our model's competitive edge, outpacing existing methodologies in R-Precision and diversity, particularly with intricate textual narratives.

\section*{Acknowledgement}
This project is supported by 
the Ministry of Education, Singapore, 
under its Academic Research Fund Tier 2 
(Award Number: MOE-T2EP20122-0006).

{\small
\bibliographystyle{ieee_fullname}
\bibliography{egbib}
}

\end{document}